\documentclass{IOS-Book-Article}

\newcommand{\ben}{\begin{enumerate}}
\newcommand{\een}{\end{enumerate}}
\newcommand{\bit}{\begin{itemize}}
\newcommand{\eit}{\end{itemize}}

\newcommand{\A}{\mathcal{A}}
\newcommand{\B}{\mathcal{B}}

\newcommand{\PA}{\mathcal{P}}

\newcommand{\begintab}[1]{ \begin{tabular}{#1} }
\newcommand{\btab}[1]{\medskip \par \begintab{#1}}
\newcommand{\etab} {\end{tabular} \medskip \par \noindent }

\usepackage{microtype} 
\newcommand{\CB}{\textit{CB}}
\newcommand{\MM}{M}
\newcommand{\EM}{E}

\newcommand{\Pro}{\mathtt{Pro}}
\newcommand{\Con}{\mathtt{Con}}

\usepackage{graphicx} 
\usepackage{enumitem}
\usepackage{hyperref}
\usepackage{tikz}
\usepackage{float}
\usepackage{graphicx}
\usepackage{amsmath}
\usepackage{amsthm}
\usepackage[utf8]{inputenc}
\usepackage{wrapfig}
\usepackage{lscape}
\usepackage{rotating}
\usepackage{caption}
\usepackage{nccmath}
\usepackage{comment}
\usepackage{afterpage}
\usepackage{multicol}
\usepackage{cleveref}
\usepackage{color,soul}
\usepackage{diagbox}
\usepackage{soul}
\usetikzlibrary{shapes, arrows}

\theoremstyle{definition}
\newtheorem{definition}{Definition}[section]

\begin{document}

\begin{frontmatter}

\title{Defending the Hierarchical Result Models of Precedential Constraint}

\author[A]{\fnms{Henry} \snm{Prakken}\thanks{Corresponding Author: Henry Prakken, h.prakken@uu.nl.}} 
and
\author[B]{\fnms{Wijnand} \snm{van Woerkom}}

\runningauthor{Henry Prakken and Another Author}
\address[A]{Department of Information and Computing Sciences, Utrecht University, The Netherlands}
\address[B]{Max Planck Institute for Comparative and International Private Law, Hamburg, Germany}

\begin{abstract}
In recent years, hierarchical case-based-reasoning models of precedential constraint have been proposed. In various papers, Trevor Bench-Capon criticised these models on the grounds that they would give incorrect outcomes in some cases. In particular, the models would not account for the possibility that intermediate factors are established with different strengths by different base-level factors. In this paper we respond to these criticisms for van Woerkom's result-based hierarchical models. We argue that in some examples Bench-Capon seems to interpret intermediate factors as dimensions, and that applying van Woerkom's dimension-based version of the hierarchical result model to these examples avoids Bench-Capon's criticisms.
\end{abstract}

\end{frontmatter}

\section{Introduction}

In \cite{c+h23} and \cite{WvW23} hierarchical factor-based models of precedential constraint were proposed, extending the factor-based result and reason models of precedential constraint of \cite{hor11}. In \cite{tbc24ailj}, extending \cite{tbc23jurix}, Trevor Bench-Capon criticised these models for giving incorrect outcomes in some cases. The main point of his criticism was that the models do not account for the possibility that intermediate factors are established with different strengths by different base-level factors. In \cite{c+h23jurix}, Canavotto \& Horty responded to Bench-Capon's criticism in \cite{tbc23jurix} for their hierarchical reason model of \cite{c+h23}. In this paper we reply for the hierarchical result models of \cite{WvW23,WvW23jurix}.  Our main point will be that in the analysis of some examples Bench-Capon in fact seems to interpret some intermediate factors as dimensions, which makes the dimension-based version of the hierarchical result model proposed by  \cite{WvW23jurix} applicable to these examples (a point not made by \cite{c+h23jurix}). We show that these reinterpretations of the examples are not subject to Bench-Capon's criticisms.

%

 \section{The hierarchical factor-based result model}
 
We first summarise the factor-based flat and hierarchical result models of precedential constraint (abbreviated as RM and HRM) as presented in \cite{WvW23}.\footnote{We present the model with a minor change: only the fact situation with respect to which constraint is computed is allowed to be partial. We make this change for the simplicity of the presentation; it does not have an impact on the functionality of the model with respect to the criticisms under consideration. We make the same change to the dimensional version in Section~\ref{sec:dhrm}.} Given a set $P$ of factors, a \emph{fact situation} $F$ is a partial function $F : P \rightharpoonup \{t,f\}$, i.e., a function $F : Q \to \{t,f\}$ for some subset $Q \subseteq P$. Intuitively, $F(p) = t$ means that factor $p$ applies  in $F$ while $F(p) = f$ means that factor $p$ does not apply in $F$, and if $F$ is undefined on $p$ then this means it is (as yet) unknown whether $p$ applies in $F$ or not. A fact situation that is defined on all of $P$ is called \emph{complete}. We write $F \models p$ for $F(p) = t$ and $F \models \neg p$ for $F(p) = f$. A \emph{case} is a pair $(F,s)$ where $F$ is a complete fact situation and $s \in \{\pi,\delta\}$. Here $s$ is the outcome, or decision, of the case. The set $P$ of factors is assumed to be partitioned into two subsets $\mathtt{Pro}$ and $\mathtt{Con}$. If $p \in \mathtt{Pro}$ $(\mathtt{Con})$ then $p$ is said to support outcome $\pi$ ($\delta$). Two corresponding functions $\mathtt{Pro}, \mathtt{Con} : \{\pi,\delta\} \to 2^P$ are defined as follows: $\mathtt{Pro}(s) = \mathtt{Pro}$ if $s = \pi$ and $\mathtt{Pro}(s) = \mathtt{Con}$ if $s = \delta$ and, likewise, $\mathtt{Con}(s) = \mathtt{Con}$ if $s = \pi$ and $\mathtt{Con}(s) = \mathtt{Pro}$ if $s = \delta$. Finally, a case base $\CB$ is a set of cases. 

Now `flat' precedential constraint is defined in the result model RM as follows.

\begin{definition}\label{defRM}
Let $\CB$ be a case base for a set of factors $P$ and $s \in \{\pi,\delta\}$. We say that $\CB$ \emph{forces} the decision of a fact situation $F$ for $s$, written $\CB,F \models s$, iff there exists a case $(G,s) \in \CB$ such that 
\ben
\item for all $q \in \mathtt{Pro}(s)$: if $G \models q$ then $F \models q$, and
\item for all $q \in \mathtt{Con}(s)$: if $F \models q$ then $G \models q$
\een
\end{definition}
In words, $F$ is forced for $s$ if there exists a precedent case $(G,s)$ such that $F$ contains at least all  $\mathtt{Pro}(s)$ factors that are in $G$ and at most all $\mathtt{Con}(s)$ factors that are in $G$. In other words, $F$ is forced for $s$ if there exists a precedent case $(G,s)$ such that all differences between the two make $F$ even stronger for $s$ than $G$.

To make this hierarchical, a factor hierarchy is assumed, which is defined as follows.

\begin{definition}\label{defFH}
A \emph{factor hierarchy} is a pair $(P,H)$ with $P$ a finite set of propositional letters (factors) and $H$ a relation on $P$ satisfying
\ben
\item the transitive closure of $H$ is irreflexive;
\item $P$ contains exactly one $H$-maximal element;
\item $H$ is equal to a disjoint union $\mathtt{Pro} \cup \mathtt{Con}$ of relations $\Pro$ and $\Con$.
\een
A factor $p \in P$ is \emph{basic} if there is no $q \in P$ with $H(q,p)$, and it is \emph{abstract} otherwise. The sets of all basic and all abstract factors are denoted with $B$ and $A$, respectively. Furthermore, $\PA$ is defined as $P \cup \{\neg p \mid p \in P\}$. Likewise, 
$\B$ and $\A$ are the closures of $P$, $B$ and $A$ under negation $\neg$. Finally, $\mathtt{Pro}(p) = \{q \in F \mid \mathtt{Pro}(q,p)\}$, $\mathtt{Con}(p) = \{q \in F \mid \mathtt{Con}(q,p)\}$, $\mathtt{Pro}(\neg p) = \mathtt{Con}(p)$ and $\mathtt{Con}(\neg p) = \mathtt{Pro}(p)$. The unique $H$-maximal element is denoted by $\pi$ and its negation ($\neg \pi$) by $\delta$. 
\end{definition}

Note that case outcomes are regarded as part of the hierarchy in this model. More specifically, point 2 of Definition~\ref{defFH} requires that a factor hierarchy $(P, H)$ culminates in a single factor $\pi \in P$ which represents a decision for the plaintiff or defendant. Given a fact situation $F$, the statement that $F \models \pi$ means that $F$ was decided for the plaintiff, and $F \models \delta$ means that it was decided for the defendant. As such, a \emph{case} is now modelled simply as a complete fact situation. 

An example factor hierarchy, copied from \cite{tbc24ailj}, is displayed in Figure~\ref{figFH}, where  a $+$ denotes a  $\mathtt{Pro}$ relation while a $-$ denotes a  $\mathtt{Con}$ relation. Its unique maximal element is {\tt Ice cream}. 

The hierarchical result model HRM is then defined recursively as follows.
\begin{definition}\label{defHRM}
Let $\CB$ be a case base for a  factor hierarchy $(P,H)$ and $p \in \PA$ a factor. We say that $\CB$ \emph{forces} the decision of a fact situation $F$ for $p$, written $\CB,F \models p$, iff either: 
\bit
\item $F \models p$; or
\item $p \in \A$ and there exists a case $G \in \CB$ with $G \models p$ such that 
\ben
\item for all $q \in \mathtt{Pro}(p)$: if $G \models q$ then $\CB,F \models q$, and
\item for all $q \in \mathtt{Con}(p)$: if $\CB,F \models q$ then $G \models q$
\een
\eit
When $\CB$ is a singleton $\{G\}$ we will write $G, F \models p$ instead of $\{G\}, F \models p$.
\end{definition}
Note that for a base-level factor $p$ the statement $\CB, F \models p$ is equivalent to $F \models p$, and that the recursive clause of Definition~\ref{defHRM} reduces to the `flat' constraint model of Definition~\ref{defRM} when $\pi$ is the only abstract factor in the hierarchy.

 \section{Examples and Bench-Capon's criticisms}
 
\begin{figure}[h]
\centering
\includegraphics[scale=0.55]{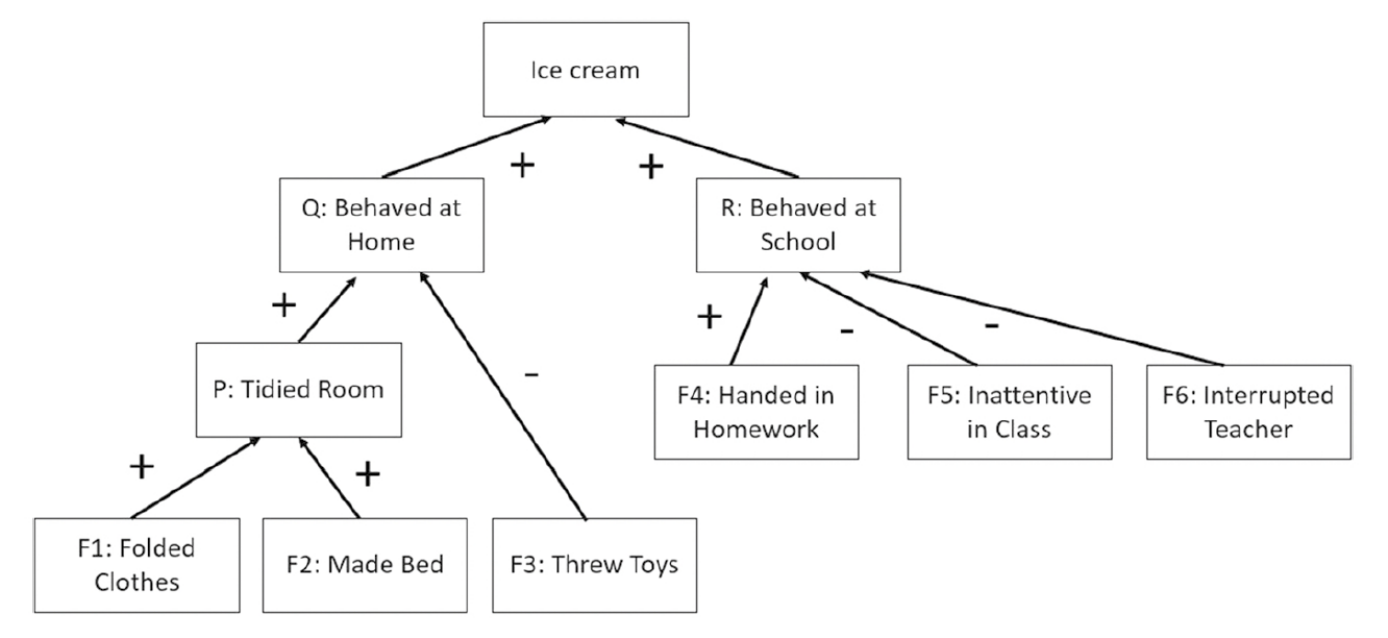}
\caption{A factor hierarchy (from \cite{tbc24ailj}, as adapted from \cite{c+h23})}\label{figFH}
\end{figure}

Two of the examples discussed by Bench-Capon are relevant for the result models. They are both discussed in \cite{c+h23}, in a (hypothetical) context of a family with parents Jack and Jo with two children Emma and Max, where the parents want to consistently decide about whether their children can have ice cream. Bench-Capon assumes the factor hierarchy depicted in Figure~\ref{figFH}, which he adapted from \cite{c+h23}. 

Throughout the paper we let `the child can (not) have ice cream' be denoted by $\pi$ ($\delta$). As a matter of fact, Bench-Capon formulates his criticisms in terms of explicit preferences between factors, which is not how the (flat and hierarchical) result models are expressed. We will see that this makes that not all of his criticisms apply to the hierarchical result model. 

\subsection{The case of MaxMonday}

In the case of \emph{MaxMonday}, $F_1$ (Max had folded his clothes) and $F_5$ (Max had been inattentive in class) were present and parent Jo decided that Max could have his ice cream.  Bench-Capon says that in a hierarchical model of precedential constraint there are two possible interpretations of this case. 
 
 \bit
  \item MMH: $Q > R$ (behaving at home is more important than behaving at school)
 \item MMF: $F_1 > F_5$ (folding one's clothes is more important than being inattentive in class)
 \eit
Note that MMH and MMF correspond to the hierarchical and flat constraint model interpretations, respectively. According to Bench-Capon, the choice between these two interpretations is not obvious. It is worthwhile quoting him in full.
\begin{quote}
Jo may give clues as to her interpretation. If she says \emph{ok, you tidied your room,
which meant you behaved at home, so you can have ice cream even though you misbehaved
at school}, she is clearly thinking in hierarchical terms. If, on the other hand
she says \emph{ok, you folded your clothes so you can have ice cream even though you
were inattentive in class}, she is clearly thinking in terms of MMF. But she might say
\emph{ok, your behaved at home by folding your clothes, so you can have ice cream even
though you misbehaved at school by being inattentive}, in which case it is less clear
which constraint is intended. The ratio of cases are often not explicitly stated: they
are identified by the subsequent judges when citing them as precedents.
\end{quote}

We first note that in the model of \cite{WvW23} cases may or may not contain intermediate factors from the factor hierarchy. When cases contain them, this is intended to mean that the decision maker has determined that this intermediate factor is present in the case. This gives some flexibility to accommodate Bench-Capon's remarks. Jo's  thinking in terms of MMH can be modelled by including the intermediate factors in the case description, while her thinking in terms of MMF can be modelled by omitting the intermediate factors from the case description. In the latter case, a new case is not constrained for any decision. If, on the other hand, the intermediate factors are included in the case description, then any new case that includes both $Q$ and $R$ is constrained for the decision that the child can have ice cream. We will explain these observations in more detail below.

In our opinion, whether the outcome in the latter case is intuitively correct is relative to what is given. If the factor hierarchy is given in a way in which all base-level factors establish intermediate factors with the same strength, then there is nothing wrong with MMH. Bench-Capon's intuition can then be explained by noting that intermediate factors are not always established with the same strength. This becomes clear from his discussion of another example. 

\subsection{The case of EmmaMonday}

In the case of \emph{EmmaMonday}, $F_2$ (Emma had made her bed) and $F_6$ (Emma had interrupted her teacher) were present. According to Bench-Capon, Emma would appeal to MMH and argue that Jo's decision about Max  constrains Jack's decision  about her to the effect that she should also have her ice cream. Bench-Capon argues that Jack could instead decide that Emma will not have her ice cream on the grounds of $F_6 > F_2$, since this preference is consistent with $F_1 > F_5$. We argue that the latter decision in fact amounts to believing that intermediate factors can in different cases be established with different strengths.  Accepting both $F_1 > F_5$ and $F_6 > F_2$ in fact says that $Q$ is more strongly established by $F_1$ than by $F_2$ and that $R$ is less strongly established by $F_5$ than by $F_6$. In the next sections we will argue that the dimension-based hierarchical result model of precedential constraint proposed by  \cite{WvW23jurix} is suitable for modelling this approach and allows a modelling of the two examples that satisfies Bench-Capon's intuitions. (Bench-Capon acknowledges in \cite{tbc23jurix} that ``Thinking in terms of factors with different strengths leads to dimensions'' but ultimately rejects dimension-based  representations in favour of factor-based representations. We will discuss his reasons for this in Section~\ref{conc}.)

However, before doing so, we now show that the HRM of \cite{WvW23} does not commit to deciding in terms of intermediate factors. It seems natural to represent Max on Monday as a case $\MM$ in which $\{F_1,F_5,P,Q,\pi\}$ apply and $\{F_2, F_3, F_4, F_6, R\}$ do not apply. Furthermore, let Emma on Monday be represented as a fact situation $\EM$ in which $\{F_2,F_6\}$ apply, $\{F_1, F_3, F_4, F_5\}$ do not apply, and which is undefined on the abstract factors $\{P, Q, R, \pi\}$. Then, we have that Emma is not forced for $\pi$ by the HRM; i.e.\ $\MM, \EM \not\models \pi$. This can be seen by noting that the statement $\MM, \EM \models \pi$ recursively reduces to the statement that $\EM \models F_1$: 
\begin{align}
    && \MM, \EM \models &\ \pi \label{step1}\\
    \text{iff} && \MM, \EM \models &\ Q \label{step2}\\
    \text{iff} && \MM, \EM \models &\ P \text{ and } \MM \models \neg F_3 \label{step3} \\
    \text{iff} && \EM \models &\ F_1. \label{step4}
\end{align}
The steps above are simply an unfolding of Definition~\ref{defHRM}. For example, for the first step we note that $\EM \not\models\pi$ and so to have $\MM, \EM\models \pi$ we must instantiate $G$ in the second bullet of Definition~\ref{defHRM} with $\MM$, which is allowed because $\pi$ is abstract and $\MM \models \pi$. Furthermore, as $\MM$ satisfies only $Q$ of $\Pro(\pi) = \{Q, R\}$, while $\Con(\pi) = \emptyset$, we see that we indeed need only $\MM, \EM \models Q$. The rest of the reasoning proceeds in a similar fashion, up to the base-level factor $F_1$ where the recursion ends. As $\EM \not\models F_1$, we have thus confirmed that $\MM, \EM \not\models \pi$, which is to say that the \emph{MaxMonday} case does not force a decision to grant ice to Emma on Monday, according to the logic of the HRM. 

So in this representation the two decisions are ultimately compared in terms of the basic factors only. However, consider an alternate scenario in which we modify $\EM$ so that $P$ applies in it, i.e.\ when $\EM \models P$. We would then have $\MM, \EM \models P$, and so by reproducing the reasoning steps above up to~\eqref{step3} we see that we would have $\MM, \EM \models \pi$ in this scenario. This shows that the HRM of \cite{WvW23} does not commit to deciding in terms of intermediate factors, but that the key is whether to include a given intermediate factor in a case description or not.

Nevertheless, Bench-Capon could have argued that the second modelling still does not capture that $F_6$ is a stronger reason against allowing ice cream than $F_5$. Since in both of our modellings $R$ could simply be ignored as not being established in either of the two cases, there seems no way in the HRM to distinguish the two cases on this ground, since ignoring $R$ implies that $F_5$ and $F_6$ can also be ignored. 

However, this is different if we slightly change both fact situations by adding that both Max and Emma handed in their homework. More specifically, let $M'$ and $E'$ be fact situations which are exactly as $M$ and $E$ respectively, except that $F_4$ applies in both $M'$ and $E'$ and that $R$ applies in $M'$. In order for $M', E' \models \pi$ to hold, we must have, in addition to the above analysis in~\eqref{step1}--\eqref{step4}, that $M', E' \models R$, which would follow if we assumed that $E' \models R$. Without this assumption, we have \[M', E' \models R \quad\text{iff}\quad E' \models F_4 \text{ and } M' \models F_6.\] Therefore, since $F_6$ does not apply in $M'$, we would have $M', E' \not\models \pi$. In other words, if we add $R$ to $M'$ but not to $E'$, we see that the parents are not constrained by $M'$ to grant ice in the situation of $E'$. One way to justify this way of representing both cases is that $F_4$ is preferred over $F_5$ as a reason for $R$ but that $F_6$ is preferred over $F_4$ as a reason against $R$.  This captures Bench-Capon's intuition about the two cases. However, this analysis is still a bit ad-hoc since in the HRM there is no way to express such preferences.  To be able to do so, we have to move to the dimension-based HRM of \cite{WvW23jurix}. 


Before doing so, we respond to another objection of Bench-Capon, namely, that intermediate factors should not play any role at all in determining precedential constraint since in common-law systems intermediate factors would not be part of the law. We are not competent to assess whether he is right for existing common-law systems, but it is easy to imagine sensible legal systems where intermediate factors do matter for precedential constraint (likewise, \cite{c+h23jurix}). To see this, we consider another variation of our two example cases. Let $M''$ be as $M$, but where $F_1$ is the only basic factor that applies. Similarly, let $E''$ be as $E$, but where only $F_2$ applies. It is easy to see that the flat result model allows $E''$ to be distinguished from $M''$ on the ground that $E'' \not \models F_1$. By contrast, if $P$ is added to both $E''$ and $M''$, then we do have $M'', E'' \models \pi$ according to the HRM. A decison-maker who holds that Emma must be allowed to have her ice cream since, like Max, she tidied her room, does not seem irrational to us. We also note that even in a model with only basic factors the decision which factors to adopt often involves a decision at which level of abstraction the facts of a case should be described (in most AI \& law models of case-based reasoning the factors are at a higher level of abstraction than the plain facts of the case). This point is also made by~\cite{roth03,r+v04}.

We therefore agree with Canavotto \& Horty in \cite{c+h23jurix}, who remark in reply to \cite{tbc23jurix} that Bench-Capon's criticism leads to giving up two important things: that decisions can be described in terms of abstract factors and that they can be justified by stating preferences between abstract factors. However, we note that this does not respond to Bench-Capon's point that abstract factors can be established with different strengths in different cases. The above examples give realistic illustrations of this possibility and we will show in the next section  that the dimension-based hierarchical result model of \cite{WvW23jurix} can account for this while allowing for intermediate concepts.

\section{The dimension-based result model}\label{sec:dhrm}

We next summarise the dimension-based flat and hierarchical result models of precedential constraint (abbreviated as DRM and DHRM) as presented in \cite{WvW23jurix}.  A \emph{dimension} is a nonempty set. Given a finite set $D$ of dimensions, a (dimension-based) \emph{fact situation} is a partial choice function on $D$, i.e., a partial function $X: D \rightharpoonup \bigcup D$ such that $X(d) \in d$ for every $d \in D$ on which $X$ is defined. A fact situation is \emph{complete} if it is defined on all of $D$. A \emph{case} is a pair $(Y,s)$ where $s \in \{\pi,\delta\}$ and $Y$ is a complete dimension-based fact situation. Finally, each dimension $d$ is equipped with a partial order $\preceq$ on $d$, where $v \preceq v'$ intuitively means that $v'$ is at least as good for  $\pi$ as $v$ and $v$ is at least as good for $\delta$ as $v'$.

Then `flat' precedential constraint is defined for the DRM  as follows.
\begin{definition}\label{defDRM}
Let $\CB$ be a case base for a set $D$ of dimensions. We say that $\CB$ \emph{forces} the decision of a fact situation $X$ for $\pi$, written $\CB, X \models \pi$, iff there exists a case $(Y,\pi) \in \CB$ such with $Y(d) \preceq X(d)$ for all $d \in D$. Likewise, $X$ is forced for $\delta$ iff there exists a case $(Y,\pi) \in \CB$ such with $X(d) \preceq Y(d)$ for all $d \in D$.
\end{definition}

The DHRM is an extension of the DRM in which the dimensions have hierarchical structure. This structure is defined analogously to Definition~\ref{defFH}. 

\begin{definition}\label{defDH}
A \emph{dimension hierarchy} is a pair $(D,H)$ with $D$ a finite set of dimensions and $H$ a relation on $D$ satisfying
\ben
\item the transitive closure of $H$ is irreflexive;
\item $D$ contains exactly one $H$-maximal element.
\een
\end{definition}

We maintain the same notational conventions and terminology as we did with factor hierarchies. For example, a dimension is \emph{base-level} if it is $H$-minimal, and \emph{abstract} otherwise. However, in contrast to factors, dimensions are not assumed to have an inherent polarity; in other words, we do not assume that the hierarchical structure $H$ is partitioned into two $\Pro$ and $\Con$ relations.\footnote{We make this deviation from the definitions of~\cite{WvW23jurix} for the sake of simplicity.} As such, we use the notation $H(d) = \{e \in D \mid H(e, d\}$ instead of the notations $\Pro(d)$ and $\Con(d)$ to refer to the direct subordinates of $d$ in the hierarchy. As with the HRM, the case outcome is now part of the hierachy, so a case is simply defined as a complete fact situation. 


Hierarchical constraint is now defined for the DHRM as follows.
\begin{definition}\label{defDHRM}
Given a case base $\CB$ and a value $v$ in some dimension $d$, a fact situation $X$ is \emph{lower bound by} $v$ and $\CB$, written $\CB \models v \preceq X(d)$ iff:
\bit
\item $v \preceq X(d)$; or
\item $d \in A$ and there is $Y \in \CB$ satisfying $v \preceq Y(d)$ such that $\CB \models Y(e) \preceq X(e)$ holds for all $e \in H(d)$. 
\eit
The \emph{upper bound by} $v$, written $\CB \models X(d) \preceq v$, is defined analogously.
\end{definition}
Together, the lower and upper bound of a dimension $d$ for $X$ define the range of values that $d$ can have in $X$ given the case base. As stated by \cite{WvW23jurix}, the idea of the recursive clause is that there is a precedent $Y$ which forces $X$ to take a value $v$ which is at least $Y(d)$, and therefore $v \preceq X(d)$ follows by transitivity from $v \preceq Y(d) \preceq X(d)$.

\section{Applying the dimension-based hierarchical result model to the examples}

We now formalise our intuitions about the family example that intermediate factors can be established with varying strengths. For ease of explanation we use the natural numbers as values but our analysis applies in exactly the same way to any partially ordered set of values. For present purposes, all we need to represent is that $R$ is satisfied more strongly in \emph{MaxMonday} than in \emph{EmmaMonday}.

The factors are now dimensions but all dimensions except $P$, $Q$ and $R$ are kept two-valued as follows:  we assume two values 0 and 1 where $0 \preceq 1$ for dimensions that correspond to pro-$\pi$ factors (in this case $F_1, F_2, F_4$) and $1 \preceq 0$ for dimensions that correspond to pro-$\delta$ factors (in this case $F_3, F_5, F_6$). Moreover, $P$, $Q$ and $R$ can have any natural number as value, where $v \preceq v'$ iff $v \leq v'$. Their hierarchical structure remains as in Figure~\ref{figFH}.

We now model the intuition that making one's bed ($F_2$) makes the room more tidied ($P$) than folding one's clothes ($F_1$), and that interrupting the teacher ($F_6$) is worse behaviour at school than being inattentive in class ($F_5$). Moreover, the more the room is tidied, the better the behaviour at home ($Q$). This is modelled by assuming fact situations $M$ and $E$ which assign $0$ to all dimensions except:
\btab{ll}
$M$: & $F_1 = 1, F_5 = 1,P = 2,  Q = 2, R = 3, \pi = 1$\\
$E$: & $F_2 = 1,F_6 = 1,P = 3,  Q = 3, R = {?}, \pi = {?}$
\etab
Note that we leave $E$ undefined on $R$ and $\pi$. The question is now whether the case $M$ forces a decision to grant ice cream to Emma according to the logic of the DHRM. Formally, this comes down to whether there is a lower bound $M \models 1 \preceq E(\pi)$. We can see as follows that this is not the case. Intuitively, the point is that although Emma behaved better at home than Max, she behaved worse at school than Max and the latter is a relevant difference that allows the parents to deny Emma her ice cream. Let us see how this follows formally, by unfolding Definition~\ref{defDHRM} with respect to the statement $M \models 1 \preceq E(\pi)$. 

To have $M \models 1 \preceq E(\pi)$ (i.e., $E$ is forced to be decided for $\pi$) we must either have $1 \preceq E(\pi)$, which is not the case because $E$ is undefined on $\pi$, or else both
\ben
\item $M \models 2 \preceq E(Q)$; and
\item $M \models 3 \preceq E(R)$.
\een
The first we indeed have, since $2 \preceq 3 = E(Q)$, so let us consider the condition that $M \models 3 \preceq E(R)$. Successively applying Definition~\ref{defDHRM} and simplifying, we get:
\begin{align}
    && M \models &\ 3 \preceq E(R) \\
    \text{iff} &&&\ 3 \leq E(R), \text{ or} \\
    && M \models &\ M(F_4) \preceq E(F_4), \text{ and} \label{conj1} \\
    && M \models &\ M(F_5) \preceq E(F_5), \text{ and} \label{conj2} \\
    && M \models &\ M(F_6) \preceq E(F_6) \label{conj3} \\
    \text{iff} &&&\ 0 \preceq 1 \text{ (as values in $F_4$)}, \text{ and} \label{conj4}\\
    &&&\ 1 \preceq 0 \text{ (as values in $F_5$)}, \text{ and} \label{conj5}\\
    &&&\ 0 \preceq 1 \text{ (as values in $F_6$)}. \label{conj6}
\end{align}
In the first step, we must either have $3 \leq E(R)$, which we do not have since $E$ is undefined on $R$, or that we can instantiate $Y$ in the second bullet of Definition~\ref{defDHRM} by $M$, resulting in \eqref{conj1}--\eqref{conj3}. As $F_4$, $F_5$, and $F_6$ are base-level dimensions, on all of which $E$ is defined, these statements respectively simplify to \eqref{conj4}--\eqref{conj6}. Note that the last of these conjuncts, \eqref{conj6}, does not hold: $1 \preceq 0$ in $F_6$ by definition, because interrupting the teacher is a con-$\pi$ factor. In other words, $E$ can be distinguished on $F_6$ with respect to $M$, and so $M$ does not constrain $E$ to take a value of at least $3$ on $R$. We have thus verified that $M \not \models 1 \preceq E(\pi)$ according to the logic of the DHRM. This means Emma can be denied her ice cream without violating precedential constraint. 


We now vary the example in a way that makes Emma's case lower bounded by the decision that she is allowed to have ice cream. More specifically, we consider a fact situation $E'$ which is as $E$ except that we replace $F_6$ by $F_5$, so $E'(F_5) = 1$ and $E'(F_6) = 0$. Because $M$ and $E'$ have the same values on the subordinate dimensions $F_4, F_5, F_6$ of $R$, we now do have the lower bound $M \models 3 \preceq E'(R)$. Moreover, we also have $M \models M(Q) \preceq E'(Q)$ because $M(Q) = 2 \leq E'(Q) = 3$. Putting these together, we now do have that $M \models 1 \preceq E'(\pi)$; so the case of Max now forces the decision that Emma is allowed to have ice cream. 


%


Finally, we note that not only intermediate factors but also base-level factors can be established with different strengths by different facts (extraneous to the model), so that, strictly, speaking, the criticism of \cite{tbc24ailj} also applies to the model that is in \cite{tbc24ailj} taken to be a correct standard, namely, the `flat' result model of \cite{hor11}. In the above family example we may, for instance, have that a child was more or less attentive in class, or that the child threw fewer or more toys or did so more or less often or more or less violently. And in the often modelled US trade-secrets domain \cite{r+a87,gra17}, for example, the strength with which the factor  \emph{Bribe-Employee} is established may depend on the precise nature of the ``questionable means'' through which ``defendant may have acquired plaintiff's information'' \cite[p.\ 242]{ale97}.

\section{Conclusion}\label{conc}

In this paper we have responded to Trevor Bench-Capon's criticism in \cite{tbc23jurix,tbc24ailj}  of hierarchical result models of precedential constraint. We have argued that his point that the models do not account for the possibility that intermediate factors are established with different strengths by different base-level factors does not apply to \cite{WvW23jurix}'s dimension-based version. We have shown that in the analysis of some examples Bench-Capon seems to interpret some intermediate factors as dimensions, and that applying \cite{WvW23jurix}'s hierarchical result model to these examples avoids Bench-Capon's criticisms. In particular, we have shown that in the factor-based hierarchical result model there is more flexibility than Bench-Capon suggests, while in the dimension-based variant Bench-Capon's intuitions can be accommodated without giving up intermediate factors as a component of a model of precedential constraint. Whether the DHRM correctly models existing common law is beyond our expertise but we believe that it is correct as a model of rational case-based decision-making for any legal system that regards intermediate concepts as relevant for such decisions.  Moreover, if intermediate concepts are applied to explain or justify (instead of constrain) outcomes (a main motivation of \cite{WvW23,WvW23jurix}, and as argued for by \cite{tbc24ailj,bench-capon2023RoleIntermediate}), then it makes sense to demand that those intermediate concepts are themselves consistently applied. The hierarchical models we discussed define the type of constraints required to reason about such consistency.

It is interesting to note that Bench-Capon in \cite{tbc23jurix} also discusses dimensions. He even says ``Thinking about factors with different strengths leads to dimensions''. But he then (referring to \cite{b+a21icail}) continues by proposing that in theories of precedential constraint dimension values should be transformed to factors. A possible drawback of this approach is that it requires to define  cut-off points of dimensions to express which values support one side or another. As argued in \cite{hp21ailj} such cut-off points are in practice not always easy to identify.

In future work it would be interesting to investigate whether Bench-Capon's criticism can for \cite{c+h23}'s hierarchical reason model of precedential constraint be met in similar ways as for the result model. This would require the formulation of a dimension-based version of the hierarchical reason model, since such a model does not yet exist. In the present paper we focused on the result model since Van Woerkom has in \cite{WvW23,WvW25ailj} argued that the result model is better than the reason model suited for  explaining the decisions of data-driven machine-learning systems, which was a main focus of his work on precedential constraint. 

\bibliographystyle{plain}

\bibliography{argument}

\end{document}